\documentclass[10pt,twocolumn,letterpaper,sigchi]{article}

\usepackage[pagenumbers]{mystyle}

\usepackage[dvipsnames]{xcolor}

\newcommand{\TODO}[1]{\textbf{\color{red}[TODO: #1]}}
\usepackage{gensymb}

\definecolor{myblue}{rgb}{0.21,0.49,0.76}
\usepackage[pagebackref,breaklinks,colorlinks,citecolor=myblue]{hyperref}
\usepackage{comment}
\usepackage{hyperref}

\title{360° Volumetric Portrait Avatar}

\author{Jalees Nehvi$^{1}$\\
\and
Berna Kabadayi$^{1}$\\
\and
Julien Valentin$^{3}$\\
\and
Justus Thies$^{1,2}$\\
\and
$^{1}$Max Planck Institute for Intelligent Systems, Tübingen, Germany
\and
$^{2}$Technical University of Darmstadt
\and
$^{3}$Microsoft
}

\begin{document}
\iftrue
    \newcommand\berna[1]{\textcolor{orange}{BK: #1}}
    
    \definecolor{jtcolor}{RGB}{0,0,255}
    \newcommand\JT[1] {\emph{\textcolor{jtcolor}{JT: #1}}}

    \newcommand\Jalees[1]{\textcolor{blue}{JN: #1}}

\else 
    \newcommand\JT[1] {}
    \newcommand\TODO[1] {}
    \newcommand\BT[2] {}
    \newcommand\WZ[1] {}

    \newcommand\berna[1]{}

    \newcommand\Jalees[1]{}
\fi

\crefname{figure}{Fig.}{Figs.}
\Crefname{figure}{Fig.}{Figs.}
\crefname{section}{Sec.}{Secs.}
\Crefname{section}{Sec.}{Secs.}
\Crefname{table}{Tab.}{Tabs.}
\crefname{table}{Tab.}{Tabs.}

\renewcommand{\paragraph}[1]{\smallskip\noindent\textbf{#1}}
\twocolumn[{
\renewcommand\twocolumn[1][]{#1}
\maketitle
\begin{center}
    \centering
    \captionsetup{type=figure}
    \includegraphics[width=\linewidth, trim={0 8.1in 0 0},clip]{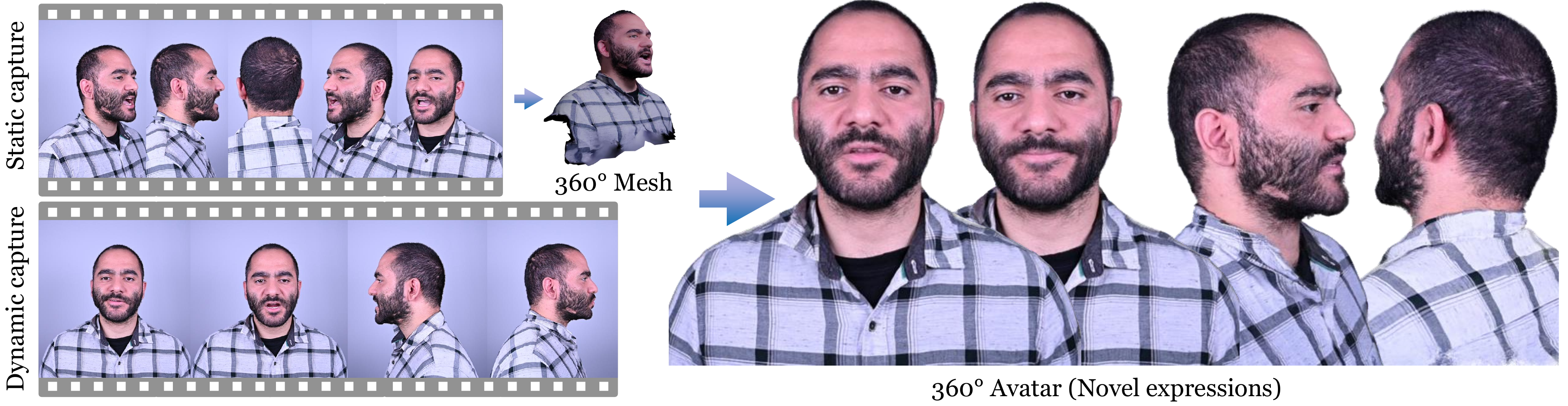} 
    \vspace{-0.4cm}
    \caption{We propose 3VP Avatar, a method to reconstruct an entire 360° portrait avatar given a monocular video capture, consisting of both static and dynamic parts (left). 
    The static capture is used to reconstruct a static mesh, which is then employed afterwards for template-based tracking of expressions, head and torso movements. We learn personalized deformation blendfields from the dynamic capture of the actor, allowing us to render a 360° avatar using novel expressions (right).}
    \label{fig:teaser}
\end{center}
\vspace{0.2cm}
    }]

\begin{abstract}
We propose 360° Volumetric Portrait (3VP) Avatar, a novel method for reconstructing 360° photo-realistic portrait avatars of human subjects solely based on monocular video inputs.
State-of-the-art monocular avatar reconstruction methods rely on stable facial performance capturing.
However, the common usage of 3DMM-based facial tracking has its limits; side-views can hardly be captured and it fails, especially, for back-views, as required inputs like facial landmarks or human parsing masks are missing.
This results in incomplete avatar reconstructions that only cover the frontal hemisphere.
In contrast to this, we propose a template-based tracking of the torso, head and facial expressions which allows us to cover the appearance of a human subject from all sides.
Thus, given a sequence of a subject that is rotating in front of a single camera, we train a neural volumetric representation based on neural radiance fields.
A key challenge to construct this representation is the modeling of appearance changes, especially, in the mouth region (i.e., lips and teeth).
We, therefore, propose a deformation-field-based blend basis which allows us to interpolate between different appearance states.
We evaluate our approach on captured real-world data and compare against state-of-the-art monocular reconstruction methods.
In contrast to those, our method is the first monocular technique that reconstructs an entire 360° avatar.
Project page: \url{https://jalees018.github.io/3VP-Avatar}.
\end{abstract}
    
\section{Introduction}
\label{sec:intro}
Creating animatable high-fidelity human avatars, finds a tremendous amount of applications in the gaming and movie industry, AR/VR, video-conferencing, \textit{etc}.
The generation of such digitized animatable content for use in industry is carried out in modern capture studios using a large, complex and expensive setup.
The pipeline for creating and animating digital human avatars captured in such studios, additionally, involves numerous complex stages and requires human experts, mostly artists, for manually designing blendshapes specific to a target avatar.
For wide masses of people that want to use their 3D appearance in a game or for immersive telepresence, those capturing procedures are not possible.
Personalized avatars need to be reconstructed with as little data as possible, with sensors that are accessible to everyone, such as a smartphone camera or a webcam.

The advent of neural volumetric rendering approaches, like Neural Radiance Fields (NeRF) \cite{mildenhall2021nerf} have revolutionized the field of 3D Computer Vision.
Given a set of RGB images and corresponding camera poses, NeRF~\cite{mildenhall2021nerf} is able to learn the volumetric radiance and density of a scene, which enables us to synthesize novel scene views at test time based on volumetric rendering.
The primary advantage offered by such methods include their implicit nature with a good gradient propagation and an implicit allocation of network capacities to local details.
Given these advantages, these approaches have been extended for the creation of animatable virtual human avatars as well.
Previous works like NeRFace~\cite{Gafni_2021_CVPR}, IMAvatar~\cite{Zheng:CVPR:2022} and INSTA~\cite{zielonka2023instant} synthesize avatars using 3DMM-based tracking of monocular videos.
NeRFace directly conditions the NeRF network on per-frame tracked expressions and jaw poses which makes it unable to generalize to unseen expressions and poses.
In contrast, IMAvatar and INSTA have a coarse deformation field obtained from a 3D morphable model (3DMM). They learn the pose dependent effects and explicitly represent the head pose and facial expressions obtained from a face tracker.
Since such methods rely heavily on the face tracker which fails to capture the non-frontal views and, particularly, back-views, one cannot obtain a full 3D avatar.
Other methods like NHA~\cite{grassal2022neural} are based on explicit mesh geometry and texture, which fail to represent thin structures like hair, \textit{etc}.
More recently, methods like VolTeMorph~\cite{garbin2022voltemorph} are able to extrapolate well to unseen poses and expressions and provide a fine-grained control but these methods do not model pose-dependent effects like wrinkles, and are trained on multi-view data.
Works like \cite{Lombardi_2019} \cite{lombardi2021mixture}, \cite{phonescanavatar} and \cite{ma2021pixel} also deploy multi-view setups to generate highly photo-realistic avatars and rely on high-quality tracking.

Our goal is to reconstruct an animatable photo-realistic avatar of a human portrait including head and torso from monocular video sequences which can be viewed from 360°.
We capture pose-dependent deformations from the captured sequences using an implicit blendshape model which is integrated with a static NeRF.
We begin by performing a 360° static scan of a human subject using a monocular RGB camera, where we focus on capturing the head and torso regions (\textit{i.e.}, the subject rotates on a chair in front of a camera).
We train a static NeRF using this information and create a personalized template mesh which is rigged using the SMPL-X body and face model~\cite{smplx}.
This template is used for tracking and for representing the coarse deformation of the avatar with respect to the canonical static NeRF.
In order to account for the fine expressions and other pose-dependent effects, we capture a dynamic sequence exhibiting a variety of expressions.
Using this dynamic data, we optimize for a corrective deformation blendfield basis as well as their corresponding coefficients.
After learning this deformation model, we train a mapping network that allows us to apply SMPL-X jaw pose and expression parameters to regress the blendfield coefficients.
In our experiments, we show that this blendfield basis allows for interpolating between different expression-dependent appearances.
We compare our method to monocular state-of-the-art avatar reconstruction methods like FLARE~\cite{bharadwaj2023flare}, INSTA~\cite{zielonka2023instant} and PointAvatar~\cite{zheng2023pointavatar}, where we can see the benefits of our proposed method.

\begin{figure*}[t]
  \centering
  \includegraphics[width=\textwidth,trim={0 11.8cm 0 0},clip]{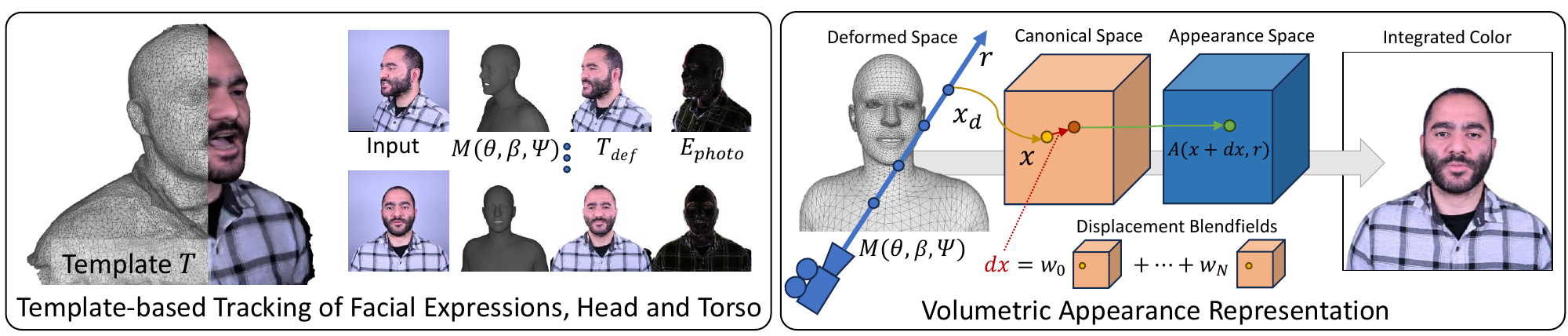}
   \caption{
    \textbf{Overview.}
    Based on monocular video inputs, we construct a template model of the subject which is used for tracking.
    It also serves as a foundation for the volumetric appearance representation that is learned conditioned on the tracked facial expression and pose parameters.
    To map samples used in the volumetric rendering from the deformed space to the appearance space, we employ a coarse deformation field and a personalized displacement field.
    As an output, we achieve 360° human portrait avatars that are controllable in terms of facial expression, pose and torso motion.
   }
   \label{fig:overviewfig}
\end{figure*}

\medskip
\noindent
In summary, we propose a method to reconstruct 360° portrait avatars from monocular inputs.
It is enabled by the following contributions:
\begin{itemize}
    \item a personalized template-based tracking of torso, head and facial expressions, using only RGB inputs including frames capturing the back-side of the human,
    \item a hybrid deformation representation, combining linear blend skinning, 3DMM-based deformations as well as personalized blendfields learned from a dynamic sequence of the subject,
    \item and a volumetric representation based on neural radiance fields which is embedded around the template mesh and is deformed with the personalized blendfields.
\end{itemize}
\section{Related Work}
\label{sec:relatedwork}


\paragraph{Monocular Head Avatar Reconstruction.} 
Creating a volumetric head avatar from a single camera is actively explored as it offers an affordable solution and is accessible to everyone~\cite{zollhoefer2018facestar}.
Especially, the advances in neural rendering~\cite{tewari2020neuralrendering,tewari2022advances} lead to the reconstruction of photo-realistic appearances.
The Neural Radiance Field (NeRF)~\cite{mildenhall2021nerf} representation is a prominent example that has been adapted to capture volumetric head avatars. 
NeRFace~\cite{Gafni_2021_CVPR} combines NeRF with a 3DMM-based face tracker~\cite{thies2016face}, which provides additional conditioning to the NeRF MLP. Thus, it models a dynamic neural radiance field and is able to produce interpolations of the training expressions and poses.
IMAvatar~\cite{Zheng:CVPR:2022} uses an implicit surface (SDF) represented by an MLP and jointly learns geometry and expression deformations.
INSTA~\cite{zielonka2023instant} relies on the coarse deformation field spanned by a 3DMM. Using Instant Neural Graphics Primitives (INGP) ~\cite{muller2022instant} to represent the radiance field, it reconstructs an avatar within a few minutes.
Similarly, FLARE~\cite{bharadwaj2023flare} produces relightable head avatars in 15 minutes by optimizing vertex displacements on top of FLAME~\cite{FLAME:SiggraphAsia2017} mesh followed by a deformation network.
It is similar to NHA~\cite{grassal2022neural} which represents the surface explicitly by a mesh, and optimizes for displacements and texture. NHA can produce complete head avatars but has missing geometric details and often produces artifacts near the ears. 
In contrast to a surface representation like a mesh, PointAvatar~\cite{zheng2023pointavatar} reconstructs a point cloud. It uses a deformation network with an additional shading network.
The point-based representation enables fast point-based rasterization and does not need a backward mapping from the deformed space to the canonical space as \cite{Zheng:CVPR:2022,zielonka2023instant}.
Note that PointAvatar~\cite{zheng2023pointavatar} performs test time optimization.
Recently, DELTA~\cite{feng2023learning} has been proposed which combines the benefits of explicit and implicit representations.
It models the skin as a mesh, and hair and accessories as a NeRF.
Other methods resemble the way classical blend shapes are employed in 3DMMs. 
Gao et al.~\cite{Gao_2022} combine multi-level hash tables with tracked expression coefficients in latent space and learn personalized NeRF-based blend shapes, which enables semantic editing of expressions. 
Besides these subject specific methods, there are techniques~\cite{wang2023styleavatar,xu2023latentavatar} that are based on generative networks like StyleGAN~\cite{Karras2019AnalyzingAI} or EG3D~\cite{Chan2021} which are trained on a large dataset of facial appearances.
Especially, Cao at al.~\cite{phonescanavatar} use a generative network trained on high-quality multi-view studio data which can be fine-tuned for a specific subject where only a video of one camera is available.
However, such high-quality training data is very expensive and not available outside large industry companies.
In contrast, our goal is to generate complete avatars with as little data as possible, only requiring a single video of a person rotating in front of a camera while showing different expressions.
%


\paragraph{Multi-view Head Avatar Reconstruction.} 
Multi-view calibrated setups are utilized to create photo-realistic, high-quality facial avatars as well as geometry and appearance priors. 
These setups employ a varying number of cameras, ranging from several~\cite{beeler2010high} to hundreds~\cite{wuu2023multiface}.
Lombardi et al.~\cite{Lombardi_2018} use a large scale multi-camera rig~\cite{wuu2023multiface}, and unwrap images to view-dependent texture maps given a tracked mesh and image. The unwrapped texture maps are averaged over all cameras.
From an average texture per frame and the tracked mesh, view-specific texture maps and the mesh are learned through an autoencoder, which is conditioned on the viewpoint, enabling one to learn view-specific texture maps.
Another mesh-based avatar method, Pixel Codec Avatars (PiCA)~\cite{ma2021pixel}, decodes only visible pixels in screen space, making facial animation efficient. 
Both Lombardi et al.~\cite{Lombardi_2018} and PiCA~\cite{ma2021pixel} produce impressive results, but they struggle in regions where tracking is difficult (i.e., mouth interior) or rendering thin structures.
Neural Volumes (NV)~\cite{Lombardi_2019} is a grid-based approach and optimizes a 3D CNN to produce a volumetric representation of the scene, but is limited in terms of resolution and memory.
MVP~\cite{lombardi2021mixture} uses a sparse data structure to tackle the shortcomings of naive grid-based approaches, such as NV~\cite{Lombardi_2019}, by minimizing the computation in empty regions and modeling the occupied regions with volumetric primitives, but requiring a high-resolution mesh.
Since these methods require days of training and expensive data captures, Cao et al.~\cite{phonescanavatar} leverage multi-view facial captures, 255 identities, and train a universal model; later, it is personalized to unseen identities leveraging universal expression codes from a phone scan.
PVA~\cite{raj2021pva} predicts volumetric avatars from a few views, aiming for generalization across identities without requiring 3D supervision.
DRAM~\cite{dram} includes a one-light-at-a-time (OLAT) prediction network, and synthesized avatars that can be relit under novel light conditions.
Extending VolTeMorph~\cite{garbin2022voltemorph}, BlendFields~\cite{kania2023blendfields} trains multiple radiance fields for each expression. It is able to generate novel expressions beyond training data by blending radiance fields based on the local volumetric changes. 
In contrast to these multi-view methods, our approach, 3VP Avatar, uses a single RGB camera, which is cost-effective and accessible to everyone and it creates an animatable head avatar, including the back and torso.
\section{Method}
Our method 3VP Avatar, reconstructs a $360\degree$ upper body avatar of a human subject, solely based on data captured with an RGB camera. 
Specifically, we assume a monocular RGB video sequence comprising a static $360\degree$ recording of a subject followed by a dynamic sequence capturing various expressions.
Our method leverages the static sequence to obtain a complete geometry and associated texture, which is subsequently used to track the dynamic part of the sequence. 
We deploy this tracked geometry and the static camera poses to reconstruct an animatable avatar by learning the 3D radiance field based on volumetric rendering. 
Our method is divided into two stages: (i) template-based tracking and (ii) volumetric modeling of appearance and deformations (see Figure~\ref{fig:overviewfig} for an overview).

\subsection{Template-based Tracking}
\label{templatetracking}
Given a monocular $360\degree$ static recording comprising the RGB images $\textit{\textbf{I}}=\{I_{i}\}_{i=1}^{N}$, where $I_{i} \in \mathbb{R}^{H \times W \times 3}$ and $N$ is the number of images,
we initially obtain the $N$ camera poses and common intrinsics along with a 3D reconstruction of the surface using structure-from-motion\footnote{\label{note1}Metashape, https://www.agisoft.com/}, which gives us a textured template mesh $\mathcal{T}$.
The reconstructed template mesh $\mathcal{T}$ along with the texture is used for tracking the dynamic part of the sequence.
Since, we are dealing with the torso region as well, we use the SMPL-X~\cite{smplx} as a 3D morphable model.
SMPL-X is an expressive 3D parametric model representing the body pose and shape, including facial expressions and hand articulations.
It is a function $M(\theta, \beta, \psi)$, parametrized by body pose $\theta$, body shape $\beta$, and facial expressions $\psi$.
We register the SMPL-X model to the template mesh $\mathcal{T}$ by utilizing detected facial keypoints and employing a 3D Chamfer distance metric:
\begin{equation}
{E}_\textrm{fit} ( \theta, \beta, \psi) = \lambda_{1}{E}_\textrm{key} + \lambda_{2}{E}_\textrm{chamf} + \lambda_{3}{E}_\textrm{reg} \, ,
\label{eq:reg}
\end{equation}
where ${E}_\textrm{key}$ minimizes the 3D distance of the facial keypoints, and ${E}_\textrm{chamf}$ calculates the chamfer distance between the vertices of $\mathcal{T}$ and SMPL-X mesh $M$.
${E}_\textrm{reg}$ is an $L_{2}$ regularizer on the body pose corresponding to the lower non-torso region of the SMPL-X model.
After aligning SMPL-X with the scanned template mesh $\mathcal{T}$, we establish correspondences based on nearest neighbor search between the template vertices and the SMPL-X mesh $M$, and transfer the linear blend skinning weights as well as the facial expression blendshape basis, which gives us a rigged and animatable scan template $\mathcal{T}_\textrm{def}(\theta, \psi)$, depending on the pose $\theta$ and expression parameters $\psi$.
Based on this rigged scan $\mathcal{T}_\textrm{def}$, we are performing non-rigid tracking.
Specifically, we use a differentiable rasterizer~\cite{Laine2020diffrast}, to perform a per-frame optimization of the jaw pose $\theta_{jaw}$, body pose $\theta_{body}$, and facial expression parameters $\psi$ based on the following energy function:
\begin{equation}
{E}_\textrm{} ( \theta, \psi) = \lambda_{1}{E}_\textrm{photo} + \lambda_{2}{E}_\textrm{land} + \lambda_{3}{E}_\textrm{temp} + \lambda_{4}{E}_\textrm{pose} + \lambda_{5}{E}_\textrm{exp}\, ,
\label{eq:two}
\end{equation}
where $\theta= \{\theta_{jaw}, \theta_{body}\}$.
Apart from the photometric term $E_\textrm{photo}$, we also add a facial landmark-based energy term $E_\textrm{land}$, temporal consistency $E_\textrm{temp}$ and regularization terms $E_\textrm{pose}$ and $E_\textrm{exp}$ on the body and jaw poses, and expressions respectively.
For optimization, we use the Adam~\cite{DBLP:journals/corr/KingmaB14} solver.
For coefficients $\lambda$ and the details of registration and tracking, please refer to Appendix~\ref{sec:RegistrationTrackingDetails}. 

\subsection{Volumetric Avatar}

\paragraph{Deformation Modeling.}
The tracked rigged meshes $\mathcal{T}_\textrm{def}(\theta_{j}, \psi_{j})$ are used for ray sampling, while the underlying SMPL-X meshes $M(\theta_{j}, \beta, \psi_{j})$ are used for modeling the dynamics of our NeRF model. Here, $j$ is the frame index which we omit in the following for better readability.
We define two types of deformations: the coarse ones provided by our tracker and an additional corrective field on top of these deformations. \\
\noindent\textit{Coarse Deformation Field.}
We construct a coarse deformation field $D_{coarse}$ using the local per-triangle deformations of the deformed SMPL-X mesh ${M}_{def}$~\cite{zielonka2023instant}. 
We employ a nearest neighbor search in the deformed space to identify the closest triangle to the sampled point.
Using the deformation gradient of this triangle with respect to the canonical SMPL-X mesh ${M}$, we transform the ray sample point from the deformed space $x_d$ to the canonical space $x$:
\begin{equation}
    x = D_{coarse}(x_d, {M}_{def}, {M}) \, .
\end{equation}
Note that for efficiency reasons, this deformation field is constructed around the relatively coarse SMPL-X model ${M}_{def}$ and not the template mesh $\mathcal{T}_{def}$, additionally, we build a bounding volume hierarchy (BVH) for a fast nearest triangle search~\cite{Clark1976HierarchicalGM}.
\\

\noindent\textit{Personalized Deformation Field.}
While the coarse deformation field models body pose and jaw motions, it is unable to model fine deformations like lip shapes and  the mouth interior behavior.
Thus, we introduce a personalized deformation field basis.
Specifically, we represent the personalized deformation field with a linear composition of a set of $K$ deformation fields $d_{i}(x)$ as:
\begin{equation}
dx(x) = d_{0}(x) + \sum_{i = 1}^{K}w_{i}d_{i}(x) \, ,
\label{eq:def_hashgrids}
\end{equation}
where $w_{i} \in \mathbb{R}$ denotes the blending coefficient.
Each of the deformation fields $d_{i}$ represents an appearance state, and is defined by its own volumetric feature grid, represented by a multi-level hash function $h_{i}(x)$, and a small MLP~\cite{muller2022instant}:
\begin{equation}
d_{i}(x) = \mathcal{D}(h_{i}(x)) \, .
\end{equation}
The network $\mathcal{D}$ takes the feature vectors obtained by querying the feature hashgrid $h_{i}$ at the canonical coordinate $x$.
$d_{0}$ represents a static deformation field, while the other deformation fields are blended with the weights $w_{i}$.
During training, the weights are optimizable per frame, and for inference, a small mapping network $\mathbb{M}$ is trained that outputs the weights based on the SMPL-X jaw pose $\theta_{jaw}$ and expression code $\psi$. 

The coarse and personalized deformation fields are used in the volumetric rendering following NeRF~\cite{mildenhall2021nerf}.
Each sample point $x_d$ on the ray $r_d$ is transformed to the appearance space $\mathcal{A}$ to obtain the density $\sigma$ and view-dependent color $c$:
\begin{equation}
\sigma, c = \mathcal{A}(x + dx(x), r_{d}) \, ,
\end{equation}
where $x$ is obtained from the coarse deformation field $x = \mathcal{D}_{coarse}(x_d)$.
The appearance $\mathcal{A}$ is represented volumetrically following INGP~\cite{muller2022instant} with an MLP which is conditioned on the ray direction $r_{d}$.
Note that we only apply the blendfield correctives $dx(x)$ to sampled points corresponding to the face region defined via the SMPL-X mesh.

\paragraph{Training.}
Given the 360° camera poses and intrinsics, the tracked meshes and the entire training image sequence, we jointly train for the canonical radiance field, the deformation field blendshapes along with the associated blending coefficients as well as the deformation field network.
We define the canonical NeRF around the registered static SMPL-X mesh.
Our objective is primarily based on the photometric loss from volumetric rendering which we define as follows:
\begin{equation}
\mathcal{L}_{color} = |\text{C} - \text{C}_{\text{GT}}|_{\epsilon} \, ,
\end{equation}
where $| .|_{\epsilon}$ is the Huber loss~\cite{Huber1964RobustEO}. We also define an additional perceptual loss~\cite{DBLP:journals/corr/SimonyanZ14a} for preserving high frequency details along with a k-fold downsampled color loss \cite{prinzler2023diner} to reduce the resulting color shifts:
\begin{equation}
\mathcal{L}_{vgg} = |{\Phi}(\text{C}) - {\Phi}(\text{C}_{\text{GT}})| \, ,
\label{eq:perceptual}
\end{equation}
\begin{equation}
\mathcal{L}_{down} = \|\mathbb{D}_{k}(\text{C}) - \mathbb{D}_{k}(\text{C}_{\text{GT}})\|_{1} \, .
\label{eq:down}
\end{equation}
A beta loss is used to enforce the opacity along a ray to sum to 1 for the foreground and 0 for the background:
\begin{equation}
\mathcal{L}_{beta} = -\sum\limits_{\Omega}\gamma\log(\gamma) \, ,
\label{eq:beta}
\end{equation}
where $\gamma$ = $\sum_{i}\alpha_{i}T_{i}$ is the opacity along a ray, $\alpha_{i} = (1 - \text{exp}(-\sigma_{i}\delta_{i}))$ with density $\sigma_{i}$ and step size $\delta_{i}$, $T_{i}$ is the transmittance and $\Omega$ denotes the ray patch. 
The overall training loss is defined as:
\begin{equation}
\mathcal{L} = \lambda_{1}\mathcal{L}_{color} + \lambda_{2}\mathcal{L}_{vgg} + \lambda_{3}\mathcal{L}_{down}+ \lambda_{4}\mathcal{L}_{beta}\, .
\label{eq:mainloss}
\end{equation}

\paragraph{Mapping Network.}
We introduce a small MLP which maps our tracked parameters, \textit{i.e.}, jaw pose $\theta_{jaw}$ and expressions $\psi$, to the learned coefficients $w$.
The mapping network $\mathbb{M}$ is only trained on the coefficients of the frontal part of the dynamic sequence, \textit{i.e.}, without side views:
\begin{equation}
w = \mathbb{M}(\theta_{jaw},\psi) \, .
\end{equation}

\paragraph{Implementation Details.}
We implement our method in PyTorch \cite{NEURIPS2019_9015}. We train our entire pipeline for 400K iterations on a single Nvidia Quadro RTX 6000 GPU.
For optimization, we use Adam \cite{DBLP:journals/corr/KingmaB14} solver with a decaying learning rate.
The training time is around 36 hours on the entire sequence.
We sample patches with size 32 during training and use $k$ = 4 for the downsampling loss (Eq.~\ref{eq:down}).
We select $\lambda_{1}$ = 1, $\lambda_{2}$ = 0, $\lambda_{3}$ = 0 and $\lambda_{4}$ = 0.1 for the first 39 epochs and set $\lambda_{1}$ = 0, $\lambda_{2}$ = 0.35, $\lambda_{3}$ = 0.035 and $\lambda_{4}$ = 0.1 for the remaining in Eq. (\ref{eq:mainloss}).

\begin{figure}[t]
\centering
    \includegraphics[width=\linewidth]{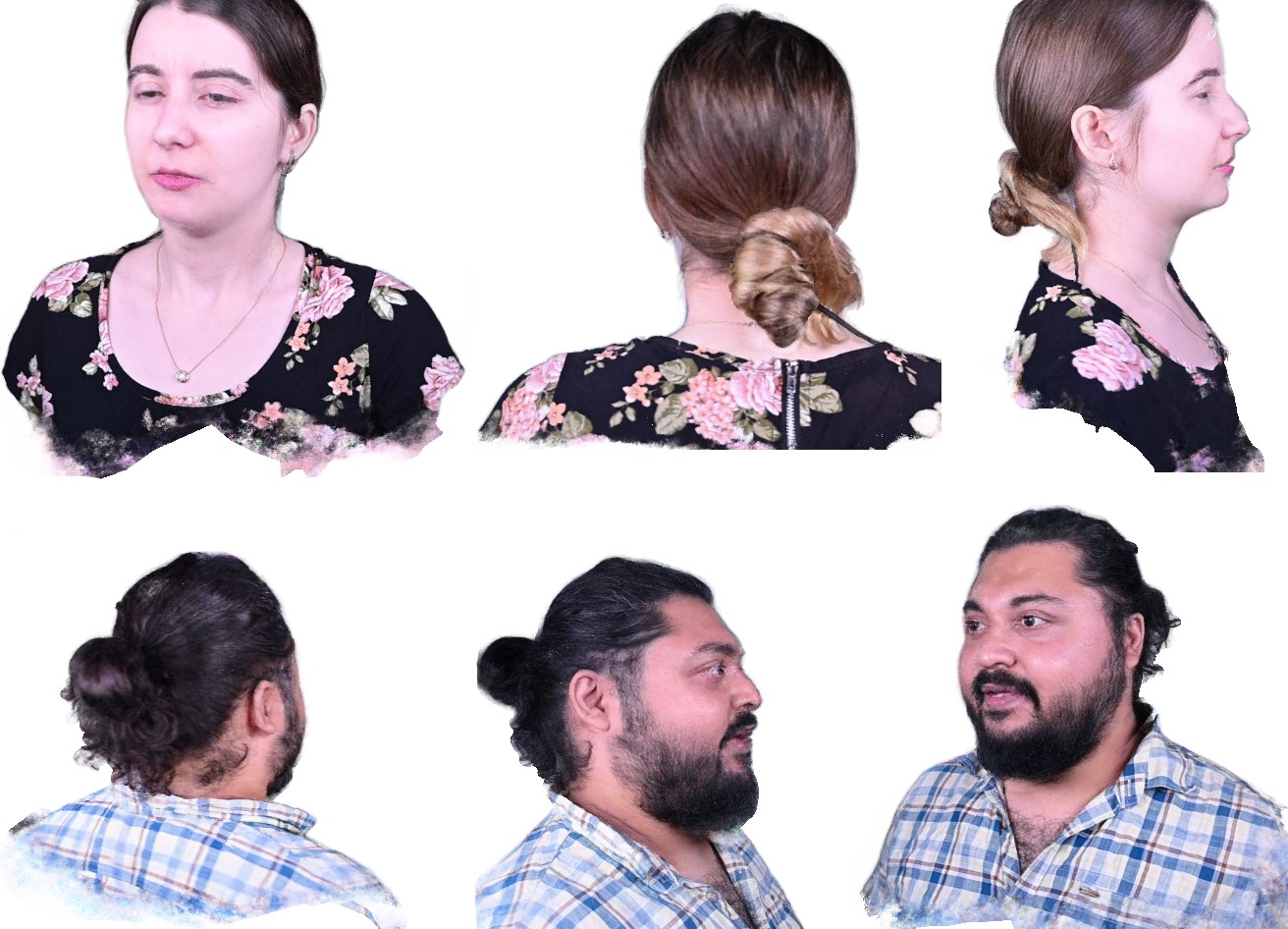} 
    \vspace{-1.05em}
     \caption{\textbf{Novel View Synthesis.} We render our 360° Avatars under novel poses and expressions from diverse novel views. Note that other methods are not able to synthesize the back view (cf. Figure~\ref{fig:sideviewcomparison}).
     }   
    \label{fig:novelviewcomp}
\end{figure}

\section{Data Capture}

We use a single Nikon Z6 II Camera for our data recordings. 
Initially, the subject is made to sit on an office chair which is then manually rotated 360 degrees to capture the full static view of the subject. The subject is expected to remain static with an open mouth while capturing the 360° sequence. This serves as our canonical state.
Afterwards, we record $K$ = 11 (Eq.~\ref{eq:def_hashgrids}) predefined expressions and visemes, each depicting an appearance state, for every subject as they rotate on the chair covering the frontal hemisphere, \textit{i.e.}, roughly 90 degrees from the front facing position to each of the sides to capture each appearance state from the frontal as well as side views.
Additionally, we also record a frontal dynamic sequence to capture the transition region between the various expression and talking states. This sequence is later used to train the mapping network as well. A similar recording with relaxed head movements is used as test data.
Our training data on average comprises of around 8000 train and 1000 test frames recorded at 25 fps in full HD resolution but cropped and resized to $512 \times 512$. Out of the roughly 8000 training frames, approximately 150 frames belong to the 360° static recording, and the remaining constitute the dynamic sequence covering the frontal hemisphere.

\section{Results}

\begin{figure}[t]
\centering
    \includegraphics[width=\linewidth]{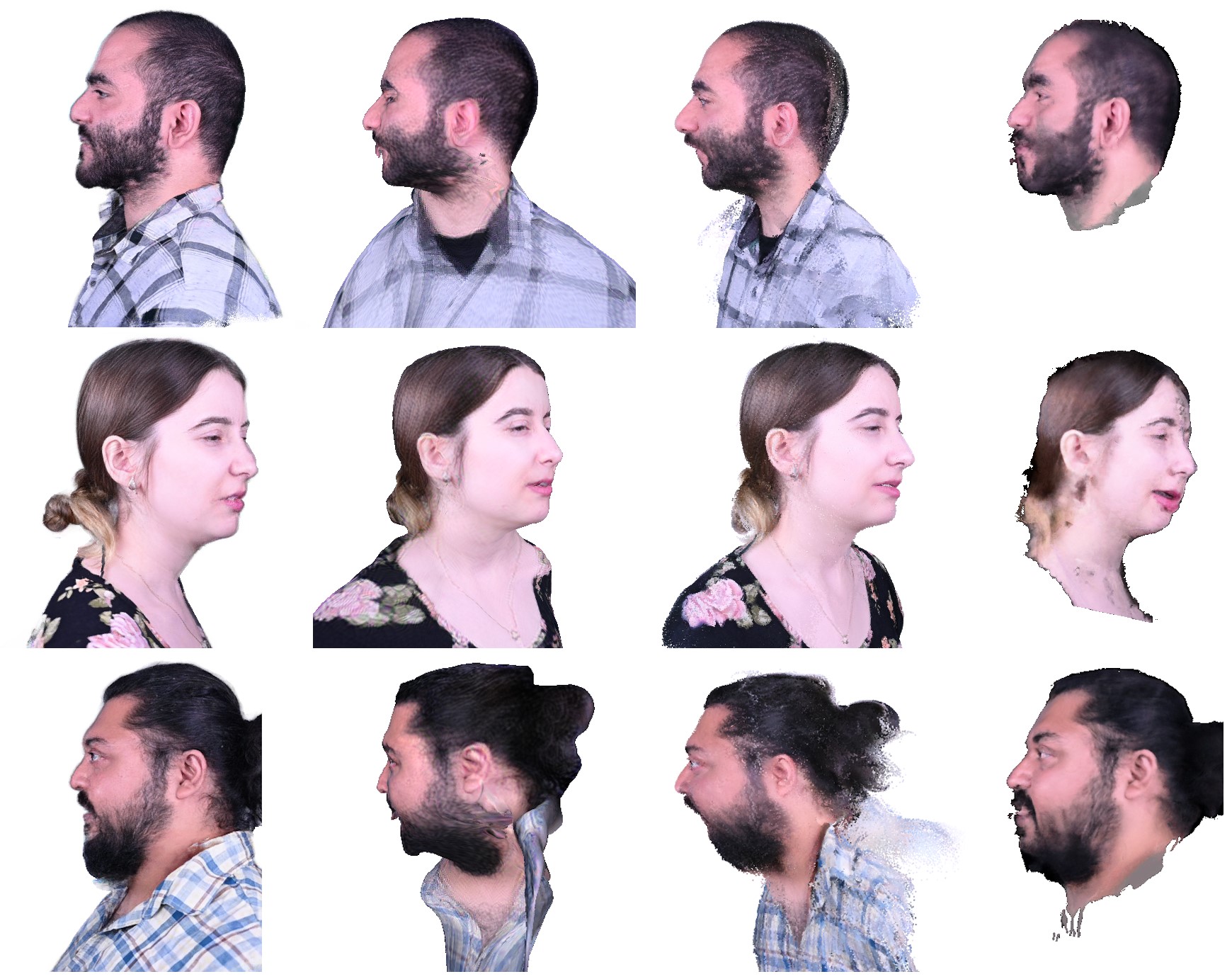} 
    \vspace{-0.55em}
    \text{\footnotesize \hspace{0.6cm} Ours  \hspace{1.0cm} FLARE \cite{bharadwaj2023flare} \hspace{0.5cm} PointAvatar \cite{zheng2023pointavatar} \hspace{0.4cm} INSTA \cite{zielonka2023instant}}
    \caption{\textbf{Side Views Comparison.} We compare the side view of our method against state-of-the-art approaches. Only our approach is able to preserve the head and torso structure of the subjects.} 
    \label{fig:sideviewcomparison}
\end{figure}

\begin{figure*}[t]
\centering
    \includegraphics[width=\linewidth]{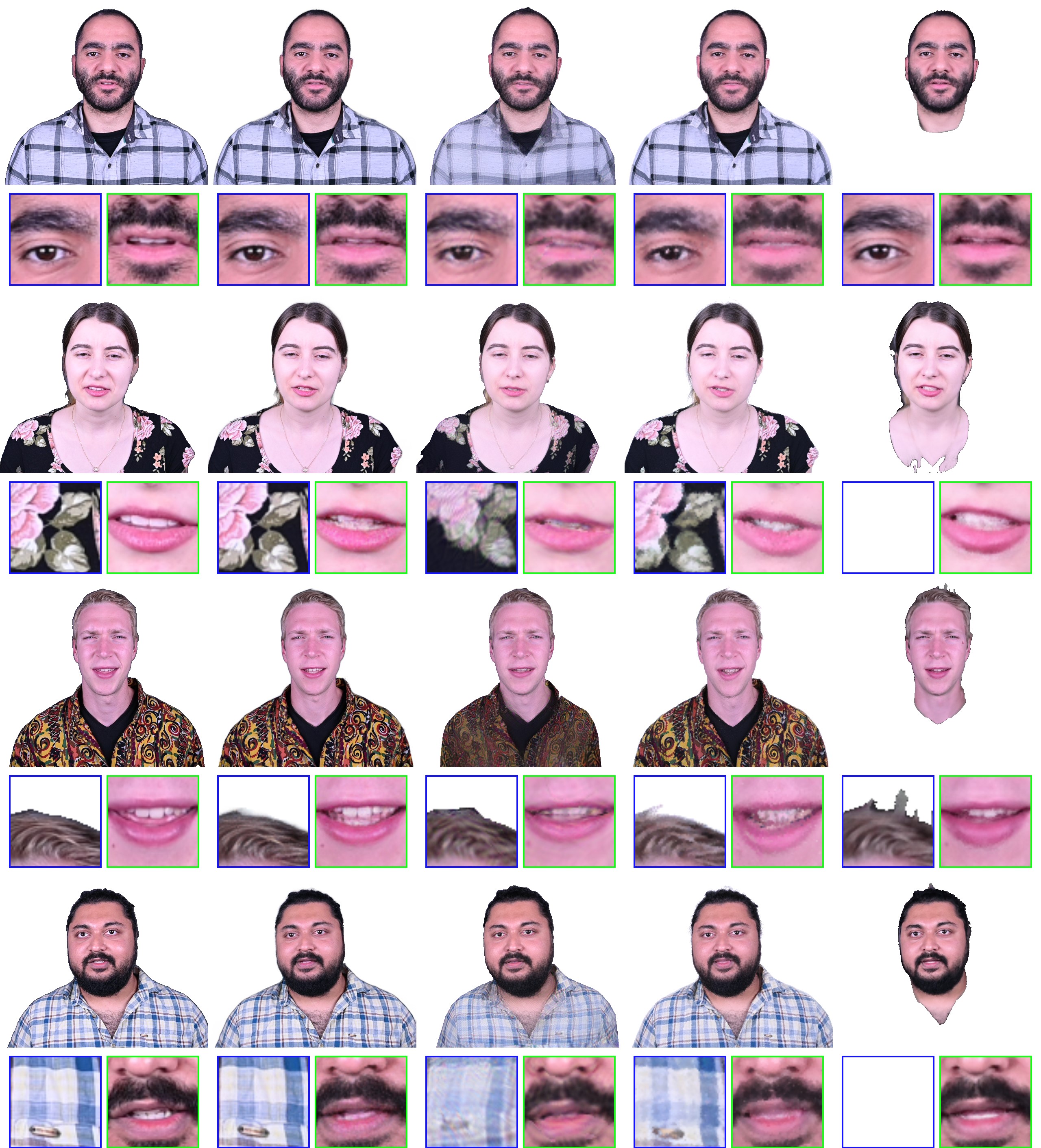} 
    \vspace{-0.75em}
    \newline
    \text{Ground Truth \hspace{1.959cm} Ours  \hspace{2.39cm} FLARE \cite{bharadwaj2023flare} \hspace{1.35cm} PointAvatar \cite{zheng2023pointavatar} \hspace{1.295cm} INSTA \cite{zielonka2023instant}}
     \caption{\textbf{Qualitative Comparison.} 
     We show a diverse set of 3VP Avatars in comparison to state-of-the-art methods. Note that our method is able to reconstruct an entire 360\degree avatar, while the other methods only reconstruct a frontal one. Nevertheless, it can be seen that the frontal appearance of our results are also more consistent and sharper, particularly in areas like the mouth interior, torso, and hair regions.
     } 
    \label{fig:comparisonsota}
\end{figure*}

\begin{figure*}[t]
\centering
    \includegraphics[width=\linewidth]{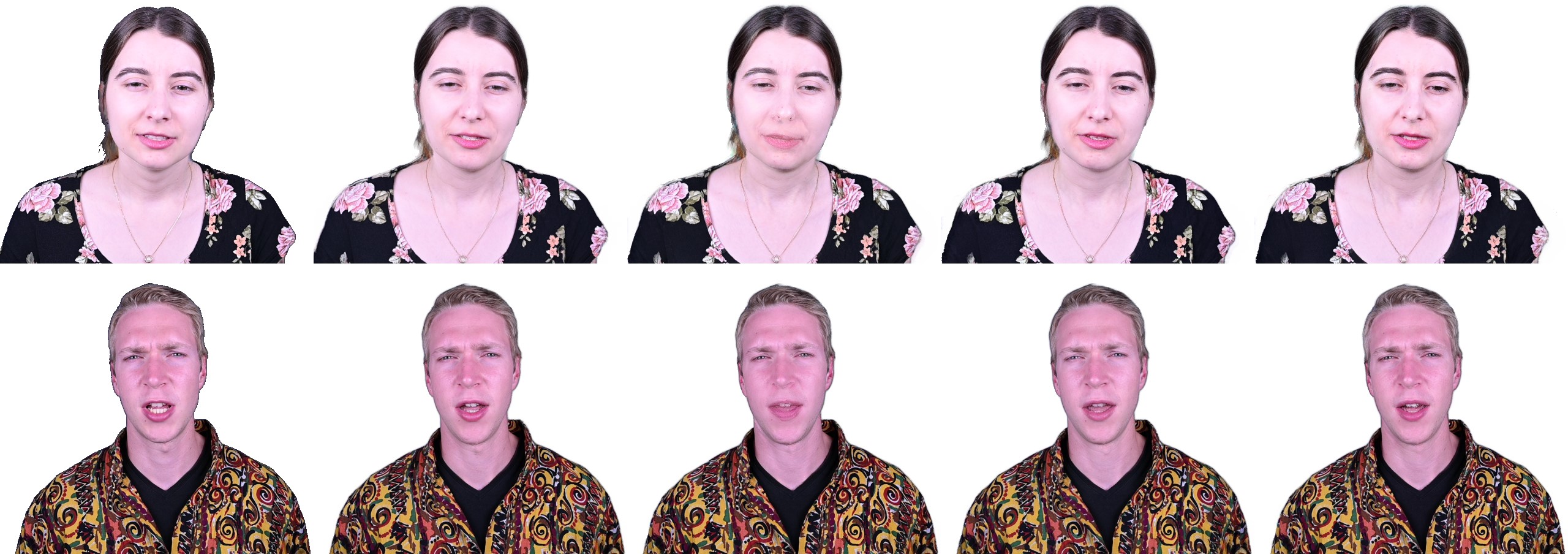} 
    \vspace{-1.35em}
    \newline
    \text{\hspace{0.7cm} Ground Truth \hspace{1.5cm} Our Method \hspace{0.95965cm} Appearance Blending \hspace{0.35cm} Direct Conditioning \hspace{0.235cm} W/o Deformation Field}
    \caption{
    \textbf{Qualitative Ablation Comparison.} We compare the visual quality of our method against different ablation experiments.
    The appearance blending experiment follows the idea of NeRFBlendshape~\cite{Gao_2022}, while the direct conditioning stems from NeRFace~\cite{Gafni_2021_CVPR}.
    } 
    \label{fig:comparisonablation}
\end{figure*}

We demonstrate our method on a dataset of 4 subjects captured using a single commodity RGB camera.
We compare our method with the state-of-the-art monocular methods: INSTA~\cite{zielonka2023instant}, PointAvatar~\cite{zheng2023pointavatar} and FLARE~\cite{bharadwaj2023flare}. 
We run the official implementations of these methods on the full dynamic part of our training set which covers the frontal hemisphere.
Through these comparisons, we demonstrate that our method, apart from reconstructing a complete 360° avatar, achieves an overall better performance on frontal test sequences as well. 
To make the comparisons fair, we compute the metrics on frontal test sequences and only in the facial region defined using GT segmentation masks obtained from INSTA.
For INSTA, we use their most recent C++ implementation with improved tracking. We train INSTA for 400K iterations. Since it only reconstructs the head region, we ignore the torso region in this case.
For PointAvatar, we train for 63 epochs on an Nvidia A100 GPU, the same settings as in the original work.
We train FLARE in the same manner as the original implementation.
%


\begin{table}[t]
    \centering
    \resizebox{\linewidth}{!}{
    \begin{tabular}{lllll}
      \toprule
      Method   & $L_{1}$ $\downarrow$ & PSNR $\uparrow$ & SSIM $\uparrow$ & LPIPS $\downarrow$ \\ 
      \midrule
      FLARE~\cite{bharadwaj2023flare}   &  0.0098  & \textbf{27.858} & 0.9437 & 0.0353 \\
      PointAvatar~\cite{zheng2023pointavatar}     &  0.0102  &  26.983 & 0.9366 & 0.0390 \\
      INSTA~\cite{zielonka2023instant}     &  0.0135  & 25.875 & 0.9333 & 0.0459 \\
      \midrule
      Our Method    &  \textbf{0.0088}  & 27.605 & \textbf{0.9470} & \textbf{0.0288} \\
      \bottomrule
    \end{tabular}}
    \caption{\textbf{Quantitative Comparison.} We compare against state-of-the-art methods: INSTA~\cite{zielonka2023instant}, PointAvatar~\cite{zheng2023pointavatar} and FLARE~\cite{bharadwaj2023flare} on the entire captured dataset comprising of 4 actor sequences and evaluate on 1000 frontal test frames.}
    \label{tab:monocularsotacomparison}
\end{table}

%
\paragraph{Metrics.}
We compute the Mean Absolute Error ($L_{1}$), the Peak Signal-to-Noise-Ratio (PSNR), the Structural Similarity Index (SSIM)~\cite{1284395}, and the Learned Perceptual Image Patch Similarity (LPIPS)~\cite{8578166} to evaluate our method quantitatively.

\paragraph{Frontal Comparison.}
Figure~\ref{fig:comparisonsota} shows the qualitative comparison of our method against state-of-the-art approaches for frontal test sequences.
Our method achieves the best quality compared to these methods.
FLARE and PointAvatar produce noisy mouth interiors. In contrast, INSTA produces a better mouth interior, but is not as sharp as our method.
This is also reflected in the quantitative evaluation in Table~\ref{tab:monocularsotacomparison}, where our method is better in all metrics except one case where FLARE has a slightly better PSNR metric.
The success of our method is mainly attributed to our stable tracking results.

\paragraph{Side Views Comparison.} 
We show in Figure~\ref{fig:sideviewcomparison} the side-view comparison of our approach against state-of-the-art methods. 
Our method clearly beats others in terms of quality as well as structure.
Even in the head region, FLARE and PointAvatar suffer from artifacts producing undesirable mesh and point cloud deformations respectively.
INSTA creates better reconstructions but produces blurry results.

\paragraph{Novel View Synthesis.}
We perform novel view synthesis from diverse camera views not seen during training.
We can see from Figure~\ref{fig:novelviewcomp} that our method produces realistic avatar reconstructions for various subjects under different novel expressions rendered from novel 360° camera poses. 

\subsection{Ablation Studies}
We conduct several ablation studies on a subset of our recorded dataset to showcase the efficacy of the different components of our pipeline. Our method is broadly divided into two main components: the tracking part and the reconstruction part. The comparisons against the state-of-the-art methods demonstrate the success of our stable tracker for tracking extreme views.
Through additional ablations, we aim to show the effectiveness of the reconstruction component of our pipeline. We show that given our tracked information, omitting the blending component or replacing it with color blending or directly conditioning on the NeRF network has a negative impact on the result quality.

\paragraph{Personalized Deformation Field.}
In this experiment, we drop the refinement component, \textit{i.e.}, the displacement field blending. We can infer from Table~\ref{tab:abaltioncomparison} that doing so results in a drop in quality in all metrics quantitatively. This is validated from Figure~\ref{fig:comparisonablation} in terms of the visual quality in the facial region as well. Thus, the introduction of the refinement component in the form of displacement field blending, on top of the coarse deformations does improve the overall quality of our facial expressions.

\paragraph{Appearance Blending.}
Similar to NeRFBlendShape~\cite{Gao_2022}, we demonstrate the effect of blending appearances instead of a deformation fields.
We follow exactly the same approach as our method, with the difference being that instead of linearly combining the deformation network predictions, the queried feature vectors corresponding to the several hash grids are blended before inputting them to the appearance network.
We can see from Table~\ref{tab:abaltioncomparison} and Figure~\ref{fig:comparisonablation} that this results in a drop in quality in the mouth region as well as the metrics. We do not directly compare against NeRFBlendShape \cite{Gao_2022} since their blending coefficients come from tracking. We observe that tracking the side views is tricky and encodes view dependency and will, therefore, make the comparison unfair.

\paragraph{Direct Conditioning.}
In this experiment, similar to NeRFace~\cite{Gafni_2021_CVPR} and INSTA~\cite{zielonka2023instant}, we condition the appearance network on tracked SMPL-X~\cite{smplx} jaw pose and expression parameters. We use the first 10 components of the expression basis along with jaw pose as conditioning to the network. Overall, the image quality is comparable to our method. Table~\ref{tab:abaltioncomparison} shows that our method achieves overall better metrics except one instance where it is marginally lower. We can see from Figure~\ref{fig:comparisonablation} that our method produces a sharper mouth interior than this NeRFace-like baseline.
\begin{table}[t]
    \centering
    \resizebox{\linewidth}{!}{
    \begin{tabular}{lllll}
      \toprule
      Experiment   & $L_{1}$ $\downarrow$ & PSNR $\uparrow$ & SSIM $\uparrow$ & LPIPS $\downarrow$ \\ 
      \midrule
      W/o Deformation Field     &  0.0099  & 27.536 & 0.9457 & 0.0698 \\
      Appearance Blending   &  0.0099  & 27.504 & 0.9421 & 0.0366 \\
      Direct Conditioning    &  0.0091  & \textbf{27.881} & 0.9488 & 0.0333 \\
      \midrule
      Ours    &  \textbf{0.0086}  & 27.880 & \textbf{0.9500} & \textbf{0.0321} \\
      \bottomrule
    \end{tabular}
    }
    \caption{\textbf{Quantitative Ablation.} We compute metrics for different ablation studies on a subset of our dataset. We observe that our method overall performs better than the ablation baselines.}
    \label{tab:abaltioncomparison}
\end{table}
\section{Limitations}
While demonstrating avatars that can be animated and viewed under 360°, there are still limitations that need to be addressed in future work.
Our tracker does not model the eye blinks and gaze.
Moreover, the training time of our method is comparably slow (36h).
We believe, that this can be reduced significantly by replacing parts of the appearance model with Gaussian splatting~\cite{kerbl3Dgaussians}, this will also remove the nearest neighbor search in deformed space.
Another limitation is that in different appearance states there are occluded regions in the mouth interior which might lead to blending artifacts for novel expressions, as the reconstruction problem of the deformation is under-constrained.
Exploring data-driven priors to model the mouth interior is an interesting avenue for future works.
\section{Conclusion}
We have presented 3VP Avatar, the first approach that generates 360° animatable avatars, including the torso, head, and back, solely based on monocular video data.
It is enabled by template-based tracking which allows us to capture the subject in 360°, eliminating the need for expensive capturing systems.
Furthermore, we propose to learn a volumetric appearance representation of the subject with a personalized deformation blendshape basis.
This deformation basis allows us to generate interpolation of different expression appearances while producing sharp results.
We believe that our work is a step towards high-quality and affordable complete digital doubles.
\\

\paragraph{Acknowledgements.}
This work was supported by Microsoft Research through its PhD Scholarship Programme.
We would like to further thank all our participants for taking part in the data recordings.
The authors thank the International Max Planck Research School for Intelligent Systems (IMPRS-IS) for supporting Jalees Nehvi and Berna Kabadayi.
{
    \small
    \bibliographystyle{ieeenat_fullname}
    \bibliography{main}
}
\appendix
\clearpage
\def\thesection{\Alph{section}}

\setcounter{section}{0}
\maketitlesupplementary
We present additional information pertaining to our method through this supplemental document. In particular, we provide further details with regards to the mesh registration step and the template-based tracker in Section~\ref{sec:RegistrationTrackingDetails} and the training procedure in Section~\ref{sec:TrainingDetails}. We describe the network architecture used in our models in Section~\ref{sec:NetworkArchitectureDetails}. We also mention some additional comparisons in Section~\ref{sec:AdditionalComparisons} and provide further ablation studies in Section~\ref{sec:AblationStudiesDetails}. Finally, we talk about the broad impact of our work in Section~\ref{sec:BroaderImpact}.
\section{Registration and Tracking Details}
\label{sec:RegistrationTrackingDetails}

\paragraph{Registration.}
\label{sec:RegistrationDetails}
We follow a three-step registration process. The first two steps are initial alignment steps that have not been discussed in the main paper. In the first step, we optimize for the scale as well as the global rotation and translation parameters of the SMPL-X~\cite{smplx} model by only using the back-projected 3D facial and body landmarks as a loss function to roughly align the SMPL-X mesh with the template scan. In the second step, we further refine the same parameters by using a Chamfer distance metric instead and finally, we optimize for all the 3DMM parameters by minimizing the objective defined in Eq.~\ref{eq:reg}, where for our experiments, we set $\lambda_{1}$ = 10, $\lambda_{2}$ = 0.01 and $\lambda_{3}$ = 1000.

\paragraph{Tracking.}
\label{sec:TrackingDetails}
In Eq.~\ref{eq:two}, we defined the objective function for our template-based tracker where we set $\lambda_{1}$ = 1.0, $\lambda_{2}$ = 0.001, $\lambda_{3}$ = 4.0, $\lambda_{4}$ = 0.1 and $\lambda_{5}$ = 0.1. We initialize the jaw and expression parameters using DECA~\cite{DECA:Siggraph2021} and regularize the jaw pose and expression code for each frame towards corresponding DECA estimates. We also strongly regularize the non-torso lower body pose to zero to prevent unwanted movement in these areas. We use a learning rate of 8e-4 which is halved every 200 iterations. To ensure smooth tracking and prevent jittering, we also apply Gaussian blurring to the rendered as well as the ground truth RGB images and define the photometric loss on these Gaussian-blurred images.

We assume that the clothing worn by the subject with regards to the torso region is rich in texture for our tracking to work properly for this region since rich textures provide sufficiently large gradients for efficiently minimizing the photometric loss in these areas.

\section{Training Details}
\label{sec:TrainingDetails}

\paragraph{View Dependency of Tracking.}
\label{sec:ViewDependencyofTracking}
We discuss a particular limitation of monocular tracking in general involving non-frontal views. Monocular tracking methods as well as monocular face reconstruction methods like DECA~\cite{DECA:Siggraph2021} are unable to provide reliable tracking information, in terms of the facial expressions and jaw pose when tracking the face from different views. This was observed in our case as well where the expression and jaw pose parameters of the 3DMM for a unique expression or viseme resulted in a range of different values that varied with the camera viewing direction. This view dependency led us to avoid using the tracked expressions and jaw pose as coefficients for our displacement blendfields. This approach can work only in cases where the face of the subject is mostly front-facing like in the case of NeRFBlendShape~\cite{Gao_2022}.

\paragraph{Sampling.}
\label{sec:SamplingDetails}
The data recorded by our setup captures different views but mostly comprises of frames from the frontal hemisphere, particularly from the front-facing direction. In order to prevent overfitting to the frontal sequence, and to equally represent the back-side during training, we follow a balanced sampling approach where we sample the static 360° views more frequently than the frontal frames of our training sequence. We sample the non-frontal hemisphere at a rate which is approximately 61 times the rate at which the frames corresponding to the frontal hemisphere expression states are sampled. In this way, we are able to reconstruct the entire avatar equally without overfitting to a particular view. Moreover, we also sample the frontal talking frames at twice the rate of the frontal hemisphere frames representing the various expression states in order to capture the transition regions effectively.

\paragraph{Data Preprocessing.}
\label{sec:DataPreprocessingDetails}
We crop and resize our recordings to 512 $\times$ 512 resolution and then run a background removal method~\cite{Qin_2020_PR} on all the frames. We sub-sample the static 360° part of the sequence and run a structure-from-motion technique, namely Metashape\footref{note1}, on the sub-sampled frames to obtain the template mesh and the camera poses. We export the 360° camera poses with intrinsics as well as the textured mesh and convert the camera extrinsics into a NeRF~\cite{mildenhall2021nerf} compatible format.

\section{Network Architecture}
\label{sec:NetworkArchitectureDetails}

\paragraph{Deformation Network.}
We use multi-resolution hash grids~\cite{muller2022instant} as encoding for our deformation blendfields as well as the appearance network.
Each feature hashgrid is represented as: $h_{i} \in \mathbb{R}^{L \times M \times F}$, where $L$ represents the number of levels, $M$ the grid size and $F$ the feature dimension per input. We choose $L$ = 16, $M$ = $2^{17}$ and $F$ = 4. We choose the coarsest and the finest resolution to be 16 and 2048 respectively. The deformation network is a 4 layer multi-layer perceptron (MLP) ~\cite{haykin1994neural} with 128 neurons and ReLU~\cite{agarap2019deep} activation.

\paragraph{Canonical NeRF Network.}
The canonical NeRF network is following the implementation of INGP~\cite{muller2022instant}.
We use a similar hash-grid as positional encoding with the same configuration of parameters as the blendfield hashgrids discussed above. The appearance network is divided into the density and color networks. It is a small MLP with 2 layers for the density branch and 3 for the color branch with both branches having a latent dimension of 64 with ReLU activation.

\paragraph{Mapping Network.}
As a mapping network from facial expression and jaw pose codes to the deformation field coefficients, we employ a 4 layer MLP with a latent feature dimension of 128 with ReLU as the activation function. We also apply positional frequency encoding to the mapping network input using 10 frequency components.

\section{Additional Comparisons}
\label{sec:AdditionalComparisons}
Apart from the baselines in the main paper, we also ran NeRFace~\cite{Gafni_2021_CVPR} and IMAvatar~\cite{Zheng:CVPR:2022}. The Face2Face \cite{thies2016face} tracker used in NeRFace failed to track the side-views resulting in broken reconstructions. In case of IMAvatar, the loss started exploding after training for around 15 epochs prompting the early stopping of the training. Due to the tracker failure issues and unfinished training, we decided not to include them in our comparisons.
\begin{figure}[t]
\centering
    \rotatebox{90}{\hspace{0.5cm} W/o Beta Loss \hspace{1.0665cm} W/o Perceptual Loss \hspace{0.45651cm} Imbalanced Sampling}
    \includegraphics[width=0.9\linewidth]{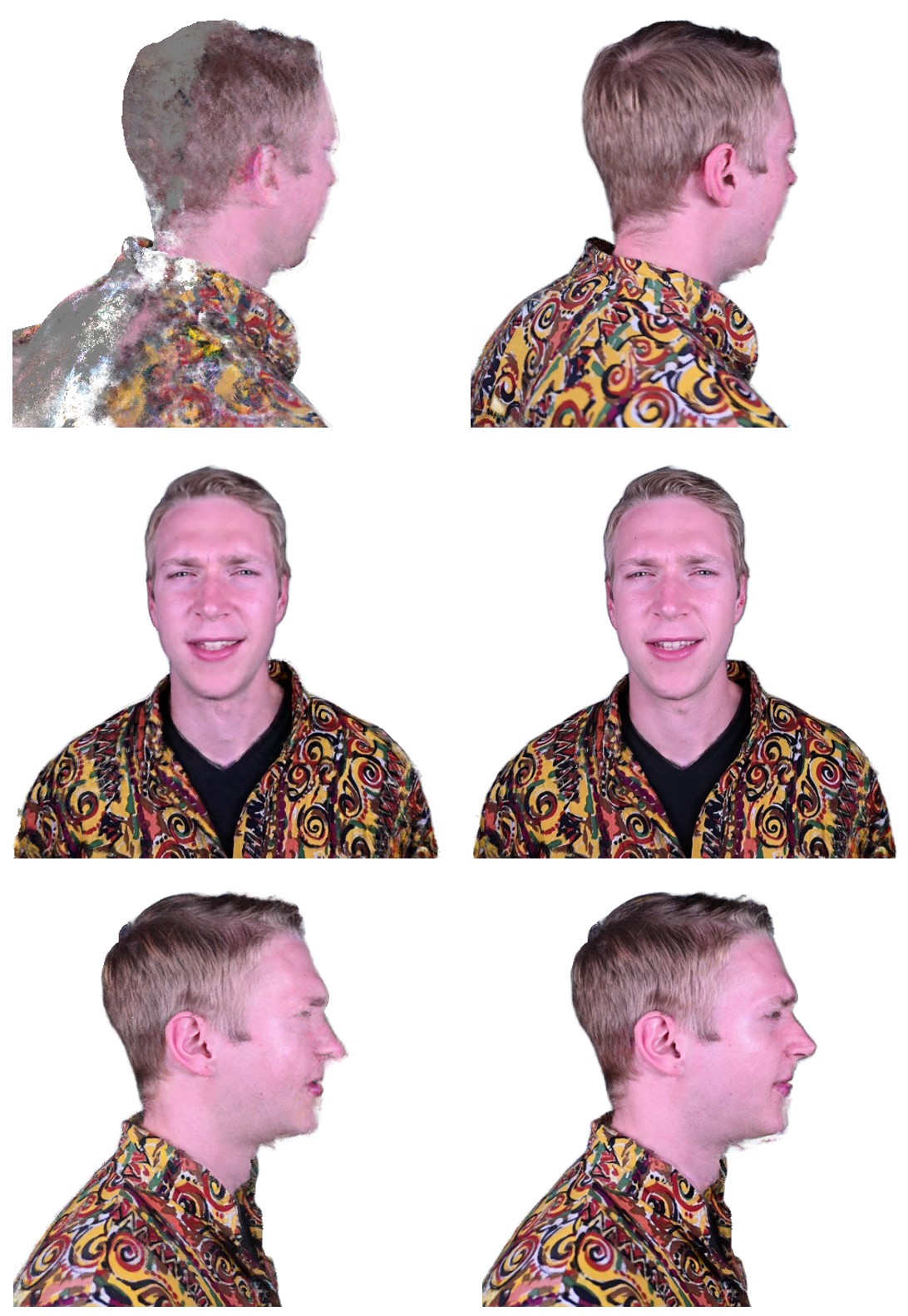}
    \text{\hspace{-0.9656cm} Ablation Experiment \hspace{1.71cm} Ours}
    \vspace{0.35em}
    \caption{We compare our method in the second column against the ablation results in the first one. Our method performs better in terms of visual quality than the ablation experiments.}
    \label{fig:4}
\end{figure}
\section{Additional Ablation Studies}
\label{sec:AblationStudiesDetails}
We conduct three additional ablation studies apart from the main paper, on the sampling procedure described in Section~\ref{sec:SamplingDetails} and the choice of used loss functions, namely, the perceptual loss (Eq.~\ref{eq:perceptual}) and the beta loss (Eq.~\ref{eq:beta}).

\paragraph{Imbalanced Sampling.}
\label{sec:RegistrationDetails}
In this study, we omit the balanced sampling and replace it with an imbalanced sampling scheme between the frontal and non-frontal views. Figure~\ref{fig:4} shows the visual result compared to our method. We can clearly observe that the image quality of the avatar rendered from the non-frontal view is strongly degraded since the back-side is under-sampled during training.

\paragraph{W/o Perceptual Loss.}
\label{sec:RegistrationDetails}
In this experiment, we omit the perceptual loss (Eq.~\ref{eq:perceptual}) term along with the downsampling loss (Eq.~\ref{eq:down}) to observe the effect on the result quality. We can deduce from Figure~\ref{fig:4} that doing so results in a blurred image compared to our method which is clearly sharper.

\paragraph{W/o Beta Loss.}
We drop the beta loss (Eq.~\ref{eq:beta}) term for foreground-background segmentation to observe its impact on the visual quality. As we can see from Figure~\ref{fig:4}, doing so results in a fuzzy appearance near the silhouette edges whereas our method produces a sharp and easily distinguishable foreground from the white background.

\section{Broader Impact}
\label{sec:BroaderImpact}

Our method is the first method that generates 360° avatars from monocular video inputs.
The recording equipment needed is a single camera or a smartphone.
This enables a broad user base to use our method, as such devices are omnipresent.
Thus, 3D avatars can be created by everyone which is essential for any immersive telepresence application in AR or VR.
Besides teleconference applications, these avatars could be used in other entertainment or e-commerce applications (\textit{e.g.}, virtual mirrors).
However, the captured 3D appearance of a person could also be misused for identity theft or other crimes.
Therefore, digital media forensics has to be explored~\cite{roessler2019faceforensics++}, to enable the detection of synthetic media.

All of our participants who volunteered for the data recordings have provided a signed consent form authorizing the use of their video recordings for our submission.

\end{document}